# LexGram
## - a practical categorial grammar formalism -

Esther König[*]
University of Stuttgart, Institute for Computational Linguistics
Azenbergstr. 12, 70174 Stuttgart, Germany
e-mail: esther@ims.uni-stuttgart.de

**Abstract**

We present the LexGram system, an amalgam of (Lambek) categorial grammar and Head Driven Phrase Structure Grammar (HPSG), and show that the grammar formalism it implements is a well-structured and useful tool for actual grammar development.

## 1 Introduction

Grammar development becomes easier if the grammar formalism which has been chosen caters for the basic needs of syntactic modelling. According to the current state of affairs, grammars for natural languages differ in the following two points from grammars for formal languages. First, the lexicon contains complex syntactic information. Second, natural language admits nonlocal syntactic dependencies.

Complex lexical information can be kept manageable if the grammar formalism includes an inheritance mechanism. For a reliable and efficient handling of nonlocal dependencies, moved and empty constituents must be official components of the grammar formalism. The existence of a representational tool which allows for relating dislocated syntactic constituents systematically with their base positions, leads to more concise grammars, i.e. modelling effort is saved. On the processing side, unnecessary struggles with getting the grammar actually running will be avoided if the conflict in information flow between the top-down propagation of a moved constituent and the bottom-up manner of building a base-generated syntactic structure, e.g. filling a subcategorization list, is taken care of by the grammar interpreter. Head-Driven Phrase Grammars (HPSG [22]) seems to be an excellent candidate to fulfill the just mentioned requirements. They are based on typed feature terms with an inheritance mechanism and treat nonlocal dependencies by the Nonlocal Feature Principle (NFP). Unfortunately, on the processing side, due to the fact that HPSG is neither a pure (phrase structure) rule-based system nor a pure lexicalist approach a genuine and efficient HPSG-interpreter seems to be still being searched for. The direct interpretation of an HPSG grammar by a successive refinement of its phrase structure schemata as it has been suggested in [5] corresponds to a top-down parser, whose deficiencies are well-known since the early age of top-down interpreted Definite Clause Grammars: A phrase structure schema may be inserted (possibly infinitly many times) although its applicability is restricted or even excluded by the input string (e.g. exploration of the adjunct scheme although no adjuncts occur in the input string or proposal of an infinite number of empty constituents by unrestricted application of the filler-head scheme.) The more sophisticated parsing methods which have been developed for grammars with a 'context-free' skeleton, i.e. with an informative phrase structure rule component, cannot reach their full efficiency since in HPSG phrase structure rules have been turned into phrase structure schemata by moving information to the lexicon. Most of the syntactic information is not located in the place where it is expected by these parsing algorithms. Parsers for lexicalized grammars, e.g. Lexicalized Tree Adjoining Grammars or categorial grammars, are not quite adequate either, since not all of the syntactic descriptions are anchored in the lexicon (i.e. filler-head structures and adjunct-head structures). This means that although HPSG is attractive for various reasons, a variant of HPSG is required for practical applications which is amenable to efficient processing in a straightforward manner.

[*]The research reported here has been funded by the Sonderforschungsbereich 340 "Sprachtheoretische Grundlagen für die Computerlinguistik", a project of the German National Science Foundation DFG.

Since HPSG is closer to a lexicalized grammar than to a rule-based grammar[1], it seems more natural to design a lexicalized version of HPSG than to compromise the essence of HPSG by transforming the phrase structure schemata back into phrase structure rules. It turns out that this is not just a matter of taste, since lexicalized grammars favor efficient processing (cf. [23], [24]) and enhance grammar design:

- The grammar interpreter can work with a restricted view on the grammar specification: Only those syntactic descriptions have to be considered which can be accessed through the words in the input string. Hence, each step of the search for an analysis is licensed by some input word. This helps to decrease the processing time in practice due to the reduced and controlled search space. In particular, the notorious problems of grammars which admit empty constituents are kept manageable, if every trace is licensed by a lexical filler category.

- Furthermore, top-down and bottom-up approaches to parsing can be joined in ideal manner. Since there is no rule component in between a goal to be derived and the lexicon, each top-down driven step can access immediately the lexicon (without going through a 'link relation', cf. figure 1) and trigger the next bottom-up step.

- Grammar modelling becomes simpler, since there is only one viewpoint on syntactic structure: from the lexemes. In non-lexicalized grammars, generalizations can only be carried out for the lexicon and the phrase structure component in separation, cf. the specification of adjunction in HPSG which is split into a phrase structure schema plus a lexical schema. In a lexicalized grammar, this collapses into one single, lexical schema.

---

[1]For a comparison of HPSG and categorial grammars, see [21]

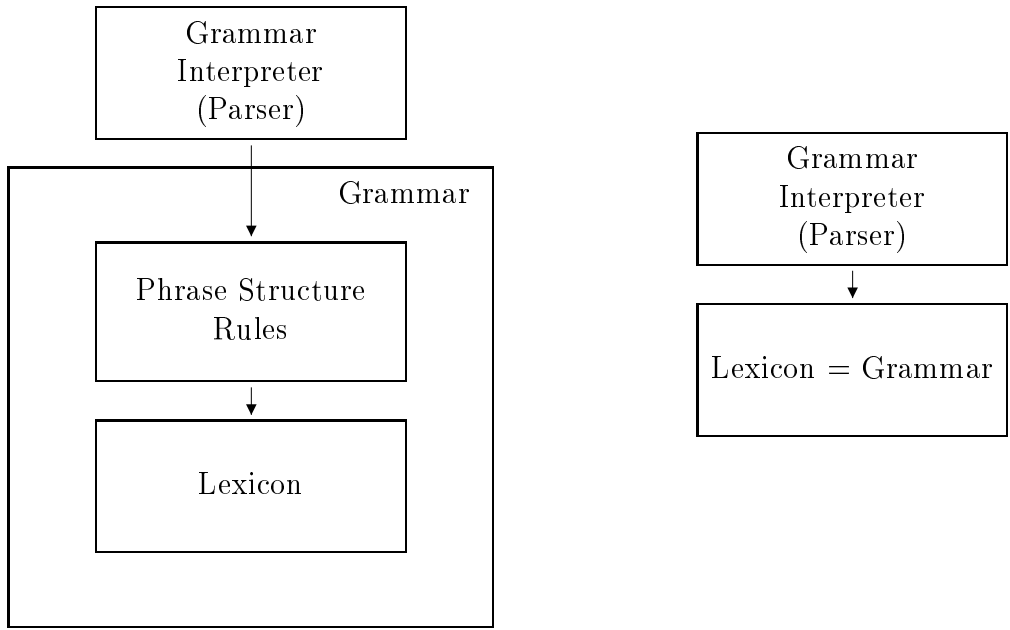

Figure 1: Architecture of phrase structure rule-based grammars vs. architecture of lexical grammars

Now, one could argue for an automatic compilation of HSPG into Lexicalized Feature-based Tree Adjoining Grammars as proposed in [20]. However, by such a compilation only the processing virtues of lexicalized grammars are gained. In order to enjoy the conceptual elegance which arises from the uniform view on grammar in lexicalized approaches, grammar should be encoded directly in a lexicalist manner. We want to propose a grammar formalism, called LexGram[2] which is derived from HPSG by lexicalizing the phrase structure principles and schemata of HPSG. At this point, the reader might object that there is nothing new about the unification-based categorial grammar which results from this enterprise. However, existing unification-based categorial grammars do not incorporate HPSG's principled way of treating nonlocal dependencies, since they take as their basis a basic categorial grammar [10], [26], [28], [2] resp. a Combinatory Categorial Grammar [27]. On the other hand, those categorial grammars which come with a logically well-defined treatment of moved constituents, the Lambek categorial grammars[3], are usually not furnished with a feature term component. Basically, the LexGram system realizes a linguistically motivated extension of the Lambek calculus[4]. LexGram has been built on top of the CUF system [3], which takes care of the handling of (recursively defined) typed feature terms.

---

[2]The LexGram system, a small grammar for German, and their documentation [13] is available via anonymous ftp.

[3]Lambek categorial grammars take the Lambek calculus [15] as their rule system. For linguistic applications and extensions see e.g. [16] and [17].

[4]The formal semantics of feature-based Lambek categorial grammars is investigated in [4].

# 2 From HPSG to categorial grammar

## 2.1 Reduction to one single phrase structure schema

Lexicalizing HPSG means to turn its phrase structure component ultimately schematic, by admitting only one single phrase structure schema. More specific postulations on the information flow in a grammar, which are not covered by that single schema, have to be expressed in the lexical signs. I.e. the phrase structure component is impoverished whereas the lexical schemata or word class descriptions are enriched.

The phrase structure component of HPSG consists essentially of the four phrase structure schemata (immediate dominance schemata) head-complement-, head-marker-, head-adjunct, head-filler structures, and of a number of principles. The major phrase structure principles are the Head Feature Principle[5]

$$
\begin{array}{c}
\left(\texttt{head : Head}\right) \\
\_\texttt{Nonlocal0} \\
\diagup \qquad \diagdown \texttt{H} \\
\_\texttt{OtherDaughter} \quad \left(\texttt{head : Head}\right) \\
\_\texttt{Nonlocal2}
\end{array}
\qquad (1)
$$

the Subcategorization Principle (here adapted to binary trees)

$$
\begin{array}{c}
\left(\texttt{subcat : SubcatRest}\right) \\
\_\texttt{Nonlocal0} \\
\texttt{C} \diagup \qquad \diagdown \texttt{H} \\
\texttt{CDtrSynsem} \quad \left(\texttt{subcat : }\left[\texttt{CDtrSynsem} \mid \texttt{SubcatRest}\right]\right) \\
\_\texttt{Nonlocal1} \qquad \_\texttt{Nonlocal2}
\end{array}
\qquad (2)
$$

and the Nonlocal Feature Principle

$$
\begin{array}{c}
\_\texttt{Local0} \\
\left(\texttt{inherited\_slash : }(S_2 \cup S_3) \setminus S_1\right) \\
\diagup \qquad \diagdown \texttt{H} \\
\_\texttt{Local1} \qquad \_\texttt{Local2} \\
\left(\texttt{inherited\_slash : } S_2\right) \quad \left(\begin{array}{ll}\texttt{inherited\_slash :} & S_3 \ \& \\ \texttt{to\_bind\_slash} & : \ S_1\end{array}\right)
\end{array}
\qquad (3)
$$

The schema for head-complement structures is the least restrictive one and closely mirrors the categorial grammar idea of a functor category 'subcategorizing' for a certain number of arguments. Hence, the head-complement schema will be the only one to survive this lexicalization enterprise. Specifiers, adjuncts, and fillers will have to become functor categories, i.e. 'heads' in a generalized sense of the word. We get rid of the head-marker schema by assuming an analysis of complementizers and determiners as functional heads, cf. [7], [19]. Instead of using a special `mod`-feature with a special treatment (bypassing the phrase structure *principles*) to express the fact that an adjunct 'subcategorizes' for the phrase it modifies,

$$\underbrace{\begin{pmatrix} \text{head} & : & (\text{mod} : \_ModifiedSynsem) \ \& \\ \text{subcat} & : & [\,] \end{pmatrix}}_{adjunct} \quad (4)$$

a categorial grammar style analysis of adjuncts `X/X`

$$\underbrace{\begin{pmatrix} \text{head} & : & \text{Head} \ \& \\ \text{subcat} & : & \left[ \begin{pmatrix} \_ModifiedSynsem \ \& \\ \text{head} & : & \text{Head} \ \& \\ \text{subcat} & : & \text{Subcat} \end{pmatrix} \mid \text{Subcat} \right] \end{pmatrix}}_{adjunct} \quad (5)$$

allows us to remove the head-adjunct schema. A treatment of fillers as heads has been proposed by [11], [6] as a transliteration of type-raised categories `X/(Y/Z)` (cf. [8], [1]) into HPSG-sign format (assuming the `subcat` value to be a list of signs):

$$\underbrace{\begin{pmatrix} \text{head} & : & X \ \& \\ \text{subcat} & : & \left[ \begin{pmatrix} Y \ \& \\ \text{inherited\_slash} & : & [\,Z, \ldots\,] \end{pmatrix} \right] \\ \text{to\_bind\_slash} & : & [\,Z\,] \end{pmatrix}}_{filler} \quad (6)$$

---

[5] We simply write the `local` and `nonlocal` feature structures as a pair, omitting the features. Feature structures are written in CUF notation, using the Prolog conventions for variables and lists.

This schema of a lexical sign means that the derivation of the complement `Y` should include a trace `Z` as one of its leaves. The trace `Z` has to be mentioned twice in order to get the desired effect in interaction with the Nonlocal Feature Principle (3). One would wish to state a more concise lexical schema for fillers, i.e.

$$\underbrace{\begin{pmatrix} \text{head} & : & X \ \& \\ \text{subcat} & : & \left[ \begin{pmatrix} Y \ \& \\ \text{slash} & : & [\,Z\,] \end{pmatrix} \right] \end{pmatrix}}_{filler} \quad (7)$$

To accomodate schema (7), the `slash`-feature has to be admitted for `synsem`-structures which occur on the `subcat` list, and the Nonlocal Feature Principle must be changed in order to retrieve the `slash` information from the complement's `synsem` value. Since no confusion can arise any longer, the feature `inherited_slash` can be omitted from the `nonlocal` structure:

$$
\begin{array}{c}
\text{\_Local0} \\
S_2 \cup S_3 \\
\overbrace{\phantom{\hspace{6cm}}}^{\text{C} \hspace{4cm} \text{H}} \\
\underset{S_1 \cup S_2}{\text{\_Local1}} \quad \underset{S_3}{\left(\, \texttt{subcat} \,:\, \left[\left(\texttt{slash} \,:\, S_1\right) \,\big|\, \_ \,\right]\,\right)}
\end{array}
\hspace{2cm} (8)
$$

Merging the Head Feature Principle (1), the Subcategorization Principle (2) and the revised Nonlocal Feature Principle (8), we get the schema in (9), which is the only ingredient of the phrase structure component of the lexicalized grammar.

$$
\begin{array}{c}
\left(\begin{array}{ll} \texttt{head} & :\ \texttt{Head \&} \\ \texttt{subcat} & :\ \texttt{SubcatRest} \end{array}\right) \\
S_2 \cup S_3 \\
\overbrace{\phantom{\hspace{10cm}}}^{\text{C} \hspace{8cm} \text{H}} \\
\underset{S_1 \cup S_2}{\begin{array}{c}\texttt{CDtrSynsem}\end{array}} \quad \underset{S_3}{\left(\begin{array}{ll} \texttt{head} & :\ \texttt{Head} \\ \texttt{subcat} & :\ \left[\left(\begin{array}{l}\texttt{CDtrSynsem \&} \\ \texttt{slash}\ :\ S_1 \end{array}\right) \,\Big|\, \texttt{SubcatRest}\right] \end{array}\right)}
\end{array}
\hspace{1cm} (9)
$$

## 2.2 Relating the single schema to a categorial grammar

In order to get access to the wealth of research concerning the logical foundations of categorial grammars, we will argue that the kernel of HPSG and a variant of the Lambek calculus are equivalent.

The `head`/`subcat` structures of HPSG are an uncurried notation for syntactic categories in a categorial grammar. The translation scheme between both representations is sketched by example 1, ignoring surface word order issues (as we will do in the remainder of this section). Essentially, /-operators at embedding level 1 mark elements of the `subcat` list, whereas /-operators at embedding level 2 correspond to the `to_bind_slash` of HPSG.

**Example 1 (Synsem's vs. categories)**

$$
\left(\begin{array}{ll} head & :\ \mathtt{X}_0 \\ subcat & :\ \left[\left(\begin{array}{ll} head & :\ \mathtt{X}_2 \\ subcat & :\ [\ ] \\ slash & :\ \left[\left(\begin{array}{ll} head & :\ \mathtt{X}_3 \\ subcat & :\ [\ ] \end{array}\right), \left(\begin{array}{ll} head & :\ \mathtt{X}_4 \\ subcat & :\ [\ ] \end{array}\right)\right] \end{array}\right), \left(\begin{array}{ll} head & :\ \mathtt{X}_1 \\ subcat & :\ [\ ] \\ slash & :\ [\ ] \end{array}\right)\right] \end{array}\right)
$$
$\Leftrightarrow (\mathtt{X}_0/\mathtt{X}_1)/((\mathtt{X}_2/\mathtt{X}_3)/\mathtt{X}_4)$ \hspace{1cm} ($\mathtt{X}_i$ atomic category symbols)

It is common knowledge that the Subcategorization Principle (2) reflects the rule of functional application $\mathtt{X/Y, Y} \to \mathtt{X}$ in categorial grammar. But what about the Nonlocal Feature Principle? Subsequently, we will show that this principle mimics the hypothetical reasoning mechanism of the Lambek calculus. To get the direct correspondence, we define a version of the phrase structure schema in (9) which is ressource-conscious with respect to the `nonlocal` information (using disjoint union ⊎ of multi-sets instead of set union). For later convenience, the string information `phonology` is added, and the, possibly dynamically generated, 'lexical' entry for the empty string is shown.

**Definition 1 (Phrase Structure Schema + Slash Termination Schema)**

*(Leftward) Phrase Structure Schema:*

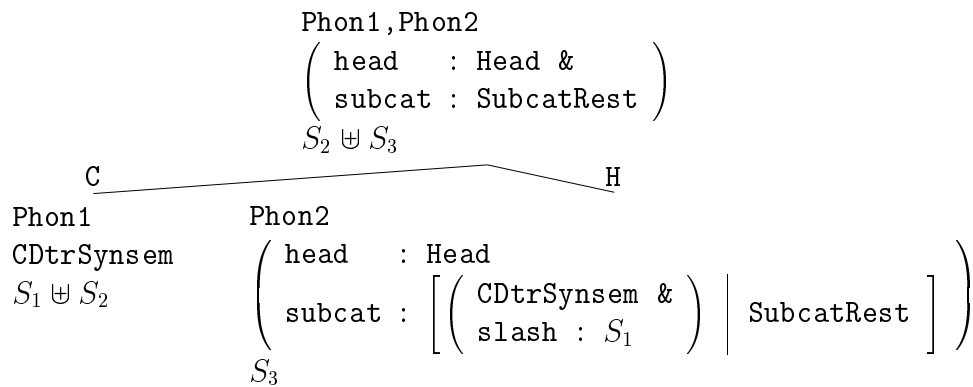

*The rightward Phrase Structure Schema is analogous with complement and head daughter interchanged.*

*Slash Termination Schema:*

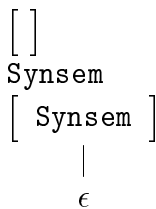

Actually, there is a close correspondence between signs and sequents (of a Natural Deduction calculus in sequent format).

**Example 2 (Signs vs. sequents)**

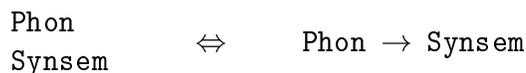

However, the `nonlocal` value does not fit into that picture. If we allow the `phon` value to be a list of words and/or `synsem`'s, Slash Termination can be incorporated into the Phrase Structure Schema. I.e. traces, which are postulated by a `to_bind_slash` are posited immediately in the complement's `phon` value instead of positing them on demand. Then the bookkeeping with respect to the `inherited_slash` can be omitted. The equivalence of definition 1 and definition 2 can be shown by inductions on the size of derivations.

**Definition 2 (non-threading Phrase Structure Schema)**

*(Leftward) Phrase Structure Schema:*

$$\begin{array}{c}
\text{Phon1, Phon2} \\
\left(\begin{array}{ll} \text{head} & : \text{Head \&} \\ \text{subcat} & : \text{SubcatRest} \end{array}\right) \\
\diagup\phantom{xx}\diagdown \\
\text{C}\phantom{}\text{H}
\end{array}$$

$\text{merge(Phon1}, S_1)$ $\phantom{xx}$ Phon2

CDtrSynsem $\phantom{xx}$ $\left(\begin{array}{ll} \text{head} & : \text{Head} \\ \text{subcat} & : \left[\left(\begin{array}{l}\text{CDtrSynsem \&} \\ \text{slash}: S_1\end{array}\right) \mid \text{SubcatRest}\right] \end{array}\right)$

The non-threading Phrase Structure Schema of definition 2 condenses an application of the rule *(/Elim)* plus $n$ applications ($n$ length of $S_1$) of the rule *(/Intro)* of a version of the Lambek calculus:

**Definition 3 (a semi-directional Lambek calculus)**

$$(ax) \quad \text{X} \rightarrow \text{X}$$

$$(/Elim) \; \frac{U_1 \rightarrow \text{Y} \quad U_2 \rightarrow \text{X/Y}}{U_1, U_2 \rightarrow \text{X}}$$

$$(/Intro) \; \frac{U_1, \text{Y}, U_2 \rightarrow \text{X}}{U_1, U_2 \rightarrow \text{X/Y}}$$

# 3 The LexGram formalism

In this section, the actual implementation of the LexGram formalism will be described.

## 3.1 Data type for lexicalized syntactic structures

In order to gain efficiency, we want the subcategorization information to reflect directly the expected surface order of the complements. In HPSG, the word order information is defined separately from the lexical subcategorization information as 'constituent order principles'. However, if word order is a part of the subcategorization information, the work of the grammar interpreter becomes much simpler because it is sufficient to use string concatenation instead of expensive permutation operations. In order to avoid confusion with the slightly different semantics of HPSG feature names, we decided to rename `head` into `root`, and `subcat` into `leaves`. The `synsem` type is replaced by the `stree` type in figure 2. Each element on the `leaves` list is annotated with its `direction` wrt. the head of the phrase. The `leaves` list induces a binary tree, where elements at the beginning of the list are closer to the head than those towards the end of the list. The set of trees[6] which correspond to uses of the `stree` type in the lexicon is described more formally in definition 4.

---

[6]The view that a LexGram grammar is a lexicon which maps strings on trees is reminiscent of Lexicalized Tree Adjoining Grammars (LTAG's) [23]. However, the two formalisms cannot be *strongly*

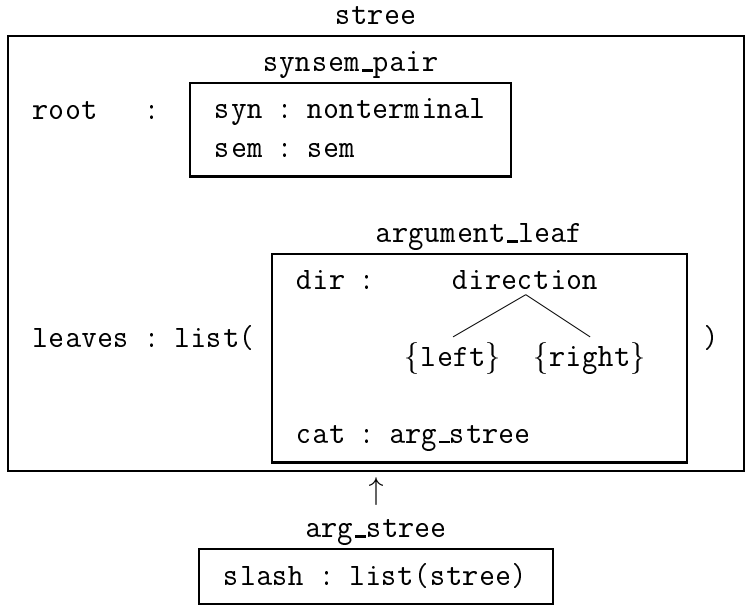

Figure 2: Data type for syntactic information

**Definition 4 (Lexical tree description)**

1. *A lexical tree description has exactly* one terminal leaf, *the* head leaf.

2. *It is a* binary *tree (except for the nontrivial subtree whose root is a preterminal node.)*

3. *All the inner nodes of the tree belong to the path between the root and the head leaf.*

**Example 3 (Lexical tree description)**

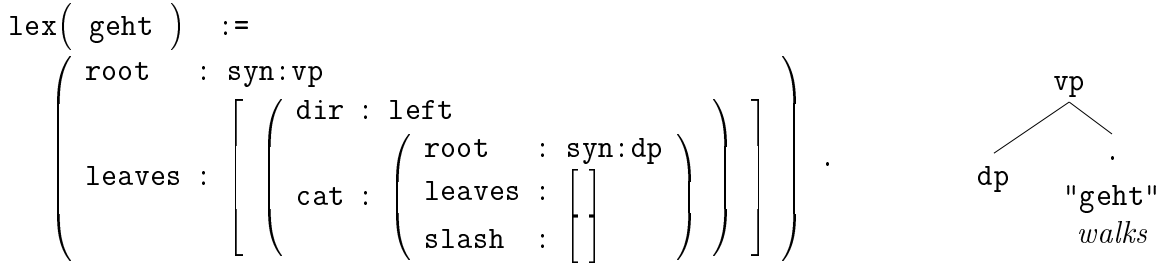

Note that open leaf nodes, i.e. expected complements, are tree descriptions themselves (of type `arg_stree`). This admits complements with a non-empty `leaves` list in order to encode control phenomena, or complements with a non-empty `slash` value for treating movement phenomena, as discussed in the section 2.

---

equivalent. On the one hand, LexGram trees are restricted to be binary trees (or flattened trees derived from binary trees). On the other hand, concerning the handling of nonlocal dependencies, it seems that the adjunction operation of TAG can only model one filler-trace relation per local tree.

*The fringe of the tree matches the input string.* *(success)*

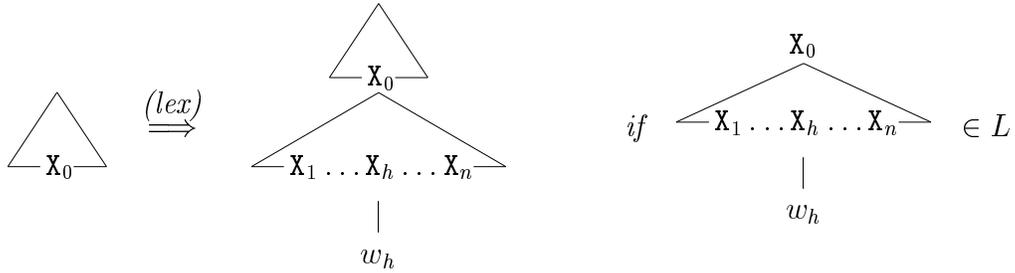

Figure 3: Graphical sketch of a head-driven parser ($L$ lexicon)

## 3.2 Grammar interpreter

The task of the grammar *interpreter* (i.e. the parser or generator) is to construct a complete syntax tree for a given string or input semantics from the partial trees specified in the grammar. Subsequently, only the parsing algorithm of the LexGram system will be presented. A generator can be spelled out along the same lines cf. König 1994 [14] and has been implemented, as well.

The parsing algorithm is an adaptation and refinement of the head-driven parser for Lambek categorial grammars (cf. [12], [9]). The basic mechanism, without traces, is sketched in figure 3 relying on the (imperfect) graphical representation of the `stree` type. To be more precise, we will give a sequent-style characterization of the full algorithm. In contrast to the Phrase Structure Schema of definition 1, the `direction` information that comes with a complement is honored by the parser. However, there is no directionality attached to the elements of the `slash` value. The list order of the `slash` is disregarded, i.e. the `slash` values are treated as multi-sets. The resulting system is in between a directional and a nondirectional Lambek calculus with the directional operators $\backslash$ and $/$ at odd levels of embedding and an undirectional operator $|$ at even levels of embedding[7], e.g. $(X_0/X_1)/((X_2|X_3)|X_4)$.

For efficiency reasons, sequences are represented as pairs of string positions[8]. Traces are posited on demand, i.e. an explicit slash threading mechanism is required. Since the `slash` value is considered as an unordered (multi-)set, an encoding by pairs of list positions is not available, only a difference-list representation `SlashIn-SlashOut`. The rule *(lex)* in figure 4 chooses a potential head from the string[9]. A trace which could serve as the head of the current phrase is taken from the incoming `slash` value by the rule *(trace)*, which guesses the insertion point of the trace in the string. The axiom

---

[7]This leads to asymmetries in the rule system (no introduction rules for $\backslash$ and $/$, no elimination rules for $|$), but this simple-minded design proved to be useful enough for writing a small, but non-trivial grammar for German.

[8]This was suggested by Jochen Dörre and implemented by Peter Krause. Interestingly, Morrill [18] introduced recently the representation of sequences as pairs of indices in a more general manner for Lambek deduction and related it to the formal semantics of Lambek systems.

[9]Sequent rules are usually read in a backward chaining manner, i.e. starting from the conclusion.

$$(\text{lex}) \quad \frac{i_1-i_2, \mathtt{X'}, i_3-i_4 \to \begin{array}{c} \mathtt{X} \\ \mathtt{Nonlocal} \end{array}}{i_1-i_2, i_2-i_3, i_3-i_4 \to \begin{array}{c} \mathtt{X}\ \& \\ \left(\begin{array}{l}\mathtt{root\ :\ R}\end{array}\right) \\ \mathtt{Nonlocal} \end{array}} \quad \begin{array}{l} w_{i_2-i_3} \in \text{input string} \wedge \\ \mathtt{lex}(w) := \left(\begin{array}{l}\mathtt{X'\ \&} \\ \mathtt{root\ :\ R}\end{array}\right) \in L \end{array}$$

$$(\text{trace}) \quad \frac{i_1-i_2, \mathtt{X'}, i_2-i_3 \to \begin{array}{c}\mathtt{X} \\ S_1, S_2-S_3\end{array}}{i_1-i_2, i_2-i_3 \to \begin{array}{c}\left(\begin{array}{l}\mathtt{X}\ \& \\ \mathtt{root\ :\ R}\end{array}\right) \\ S_1\left(\begin{array}{l}\mathtt{X'\ \&} \\ \mathtt{root\ :\ R}\end{array}\right)S_2-S_3\end{array}}$$

Figure 4: Head-driven parser: Choice of head. ($i_i$ string position; $L$ lexicon; $S_i$ slash value; $\mathtt{X}$ `stree`)

$$(\text{axiom}) \quad i_1-i_1, \mathtt{X}, i_2-i_2 \to \begin{array}{c}\mathtt{X} \\ S-S\end{array}$$

$$(PSS_/) \quad \frac{i_3-i_4 \to \left(\begin{array}{ll}\mathtt{root} & :\ \mathtt{R'\ \&} \\ \mathtt{lvs} & :\ \mathtt{L'}\end{array}\right) \quad i_1-i_2, \left(\begin{array}{ll}\mathtt{root} & :\ \mathtt{R\ \&} \\ \mathtt{lvs} & :\ \mathtt{L}\end{array}\right), i_4-i_5 \to \begin{array}{c}\mathtt{X} \\ S_3-S_4\end{array}}{i_1-i_2, \left(\begin{array}{l}\mathtt{root\ :\ R\ \&} \\ \mathtt{lvs}: \left[\left(\begin{array}{ll}\mathtt{dir} & :\ \mathtt{right\ \&} \\ \mathtt{cat} & :\ \left(\begin{array}{ll}\mathtt{root} & :\ \mathtt{R'\ \&} \\ \mathtt{lvs} & :\ \mathtt{L'\ \&} \\ \mathtt{slash} & :\ S_1\end{array}\right)\end{array}\right)\Big|L\right]\end{array}\right), i_3-i_4, i_4-i_5 \to \begin{array}{c}\mathtt{X} \\ S_2-S_4\end{array}}$$

($PSS_\backslash$) analogous to ($PSS_/$)

Figure 5: Head-driven parser: Reduction of subcategorization list

scheme *(axiom)* in figure 5 is the base of the recursion on the subcategorization list of the head in the rule *($PSS_/$)* (resp. *($PSS_\backslash$)*). Each recursion step triggers a new subproof which will be fed into the 'choice of head' rules. The `slash` values have to be represented as nested lists in order to control whether all the elements of $S_1$ are realized as traces in the derivation of the complement. A sample run of the head-driven parser is sketched in figure 6.

Empty constituents are handled safely because there are only traces available which have been licensed by words in the input string. In the current implementation of the parser, which caches the lexical lookup, parsing times are in the range of seconds. For example, it takes approx. two seconds to parse a German sentence with a dozen words which includes the following syntactic phenomena: movement of the main verb to verb first position, movement of a complement into the vorfeld, two adjectives, and a relative clause. Simultaneously to all these syntactic manipulations, a Discourse Representation Structure is built up as a feature structure.

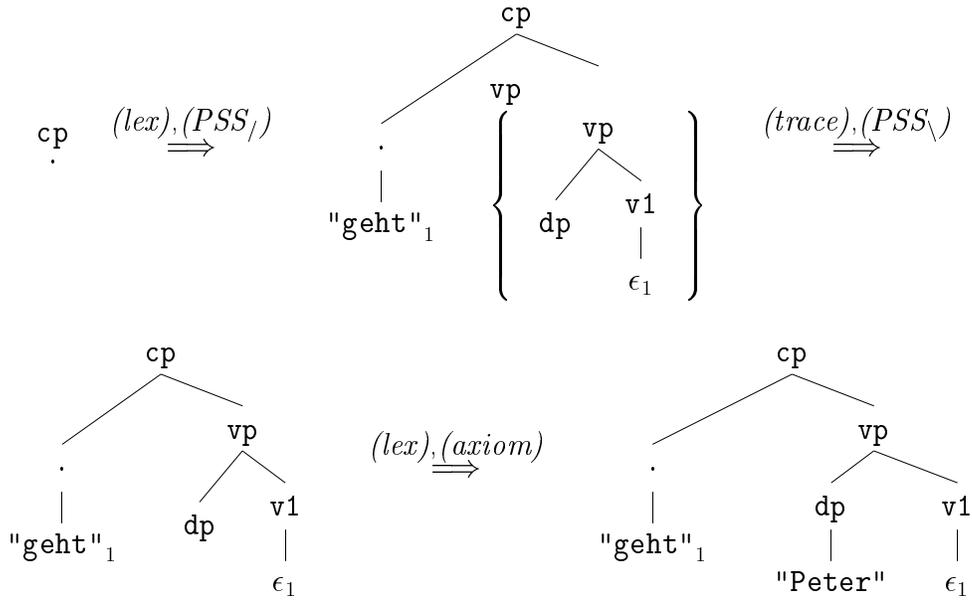

Figure 6: Sample run of the head-driven parser

## 4 Encoding a grammar

A lexicalist grammar formalism with a built-in representation of movement supports an object-centered view on grammatical descriptions: The description of the syntactic behavior of a word is composed by inheriting information along a word class hierarchy. If a constituent may be moved, an instance of a movement schema is added as one possible lexical characterization of that constituent (or its lexical head).

### 4.1 The word class hierarchy

Since all syntactic descriptions are located in the lexicon, the whole grammar can be seen as a single taxonomy of word classes whose leaves are the lexical entries, e.g. figure 7. For example, the most general syntactic structure is `syntax_tree` which helps to abstract from the realization of the `stree` data type in terms of features. `syntax_tree` is encoded as a two-place CUF sort:

$$\texttt{syntax\_tree}\bigl(\ \texttt{Root},\texttt{Leaves}\ \bigr) := \begin{pmatrix} \texttt{root} & : \texttt{Root} \ \& \\ \texttt{leaves} & : \texttt{Leaves} \end{pmatrix} \quad (10)$$

A subclass-superclass relation is represented as a CUF sort (the subclass) calling its superclass as a subgoal, i.e. `subclass := superclass`. For example, the definition of `maxproj` constrains the list of leaves to be empty:

$$\texttt{maxproj}\bigl(\ \texttt{Root}\ \bigr) := \texttt{syntax\_tree}\bigl(\ \texttt{Root},\bigl[\ \bigr]\ \bigr) \quad (11)$$

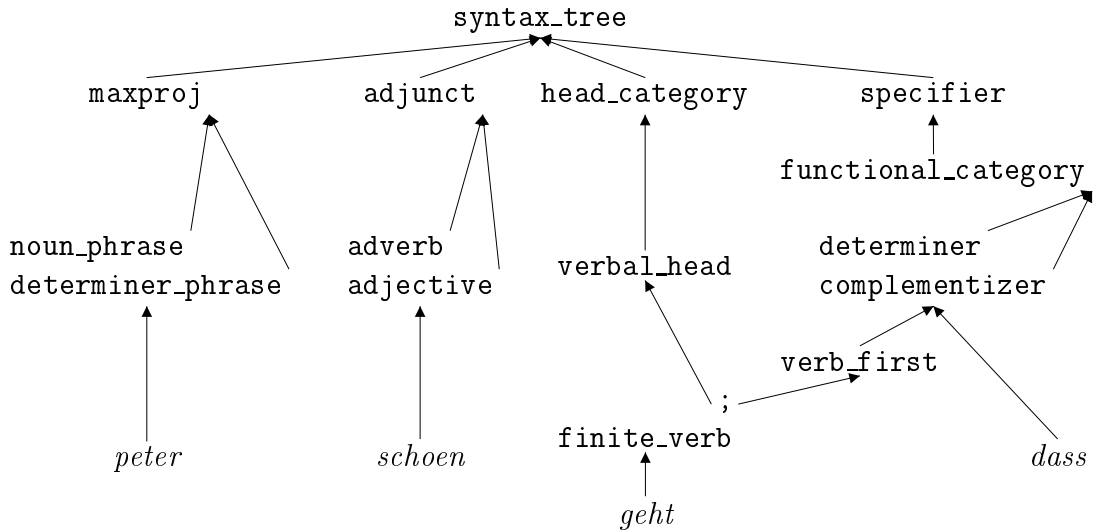

Figure 7: Part of the word class hierarchy

CUF sorts are the appropriate representation[10] because the 'expansion' of a lexical entry corresponds to a bottom-up traversal of the word class hierarchy[11], which is easily simulated by the backward chaining mechanism of proving a CUF goal.

## 4.2 Movement

Since traces are official components of the proposed grammar formalism, concise descriptions of phrase structures which involve nonlocal dependencies are feasible. Similar to the type-raising rules in categorial grammar, the movement of a phrase means that its base representation is the (single) element of the `slash` value of the (first) complement, cf. (7):

$$\begin{array}{c} \texttt{movement\_schema}\bigl(\ \texttt{Base}\ \bigr)\ := \\ \bigl(\ \texttt{leaves}\ :\ \bigl[\bigl(\ \texttt{cat:slash}\ :\ \bigl[\ \texttt{Base}\ \bigr]\ \bigr)\ |\ \_\ \bigr]\ \bigr). \end{array} \qquad (12)$$

The ability of determiner phrases to move to the `vorfeld` will be expressed by adding this alternative to the basic representation of dp's:

---

[10]Actually, CUF sorts are too expressive since no recursion is required to describe the word class hierarchy. A unification-based macro or template mechanism would be enough, cf. [25]. But for the grammar as a whole, the full expressivity of CUF sorts is needed as a declarative counterpart of what used to be called 'procedural attachments', e.g. to construct semantic representations.

[11]Note that in order to implement an HPSG faithfully, two different realizations of inheritance are required: First, inheritance which is carried out by a top-down traversal of the taxonomy of phrase structure principles and schemata in order to simulate the parsing process, i.e. a ramification in a syntax tree is obtained by specializing a phrase structure schema. Second, inheritance which is realized by a bottom-up traversal of a word class hierarchy in order to profit from lexical generalizations.

$$\begin{aligned}
\text{vorfeld}\bigl(\text{ Base }\bigr) &:= \begin{pmatrix} \text{cp\_specifier \&} \\ \text{movement\_schema}\bigl(\text{ Base }\bigr) \end{pmatrix}. \\
\text{basic\_dp} &:= \text{maxproj}\bigl(\text{ dp }\bigr). \\
\text{determiner\_phrase} &:= \text{basic\_dp}. \\
\text{determiner\_phrase} &:= \text{vorfeld}\bigl(\text{ basic\_dp }\bigr).
\end{aligned} \quad (13)$$

Compare (13) to the covert trace analysis of Pollard and Sag [22, chapter 9] where overt traces are avoided by adding alternative, shorter subcategorization frames to the word class (here the verb) whose complements can be moved. (The trace still appears on the `slash` value.) Computationally, this is more costly than the solution in (13) because e.g. three alternatives will be added to the lexicon entry of a three-place verb, which have to be explored during parsing whereas (13) adds only one alternative to a lexical entry. Furthermore, conceptually, the covert trace solution seems to be less clear: It is the dp that may move from its base position, why should the verb know about it (unless the verb puts constraints on the movement of its complement).

## 5 Conclusion

Abstract specifications in the style of HPSG have become an ideal in some groups of the computational linguistic community. However, it seems that most people who *implement* a grammar 'in the spirit of HPSG' have to compromise in order to get efficiency. In this paper, we explained one possible compromise: First, turn the phrase structure schemata and principles ultimately schematic by admitting only the head-complement schema. Second, hard-wire these principles into an efficient grammar interpreter, i.e. parser or generator.

The grammar writer is supported because the grammar interpreter incorporates a certain amount of linguistic knowledge, i.e. a general phrase structure schema including a built-in treatment of non-local dependencies. The lexicalist approach provides for a uniform view of grammar as a word class hierarchy, a clean basis for code-sharing among lexical entries. Additional benefits come from the underlying typed feature-based language. Since the signature (the type system) has to be defined before the rules, i.e. the CUF sorts, are spelled out, the writing of more transparent grammars is encouraged and errors are discovered much more often at compile time. Further, CUF sorts can serve as interfaces among the grammar modules.

So far, no specific tools for large scale grammar development and debugging are available in CUF/LexGram. A module concept is necessary to structure the grammar properly into submodules and to guarantee for data encapsulation. The tools for inspecting the data-flow in a grammar should be still improved. At compile time, a graphical representation of the word class hierarchy etc. could be extracted automatically. At run time, the user needs (more) means to visualize and to control the co-routining of subgoals.